\documentclass{article}
\input{glyphtounicode}
\usepackage{cmap}
\usepackage{microtype}
\usepackage{graphicx}
\usepackage{booktabs,tabularx,array}
\usepackage{amsmath,amssymb,amsthm,mathtools}
\usepackage{float,placeins,caption}
\usepackage{xurl}
\usepackage{hyperref}
\usepackage[preprint]{icml2027}
\hypersetup{pdftitle={Unified Dynamics of Training, Learning, and Inference in Neural Networks},pdfsubject={Neural-network training, learning, inference, feedback, and reinforcement learning}}
\graphicspath{{figures/}}
\numberwithin{equation}{section}
\DeclareMathOperator{\Cov}{Cov}
\DeclareMathOperator{\Var}{Var}
\DeclareMathOperator{\tr}{tr}

\DeclareMathOperator{\OptStep}{OptStep}
\makeatletter
\let\tli@originaltagform@\tagform@
\renewcommand{\tagform@}[1]{\tli@originaltagform@{#1}%
  \ifdefined\tli@source\textsuperscript{\normalfont[\tli@source]}\fi}
\newenvironment{sourceequation}[1]{%
  \begingroup\def\tli@source{#1}\begin{equation}%
}{\end{equation}\endgroup}
\makeatother
\newcommand{\Tmark}{\ensuremath{{}^{\scriptscriptstyle\mathrm{[T]}}}}
\newcommand{\Amark}{\ensuremath{{}^{\scriptscriptstyle\mathrm{[A]}}}}
\newcommand{\Rmark}{\ensuremath{{}^{\scriptscriptstyle\mathrm{[R]}}}}
\newcommand{\Rplusmark}{\ensuremath{{}^{\scriptscriptstyle\mathrm{[R+]}}}}
\AtBeginDocument{%
  \setlength{\abovedisplayskip}{4pt plus 2pt minus 1.5pt}%
  \setlength{\belowdisplayskip}{4pt plus 2pt minus 1.5pt}%
  \setlength{\abovedisplayshortskip}{2pt plus 1.5pt minus 1pt}%
  \setlength{\belowdisplayshortskip}{3pt plus 1.5pt minus 1pt}%
  \setlength{\jot}{2pt}%
}
\newcommand{\cU}{\mathcal U}
\newcommand{\cT}{\mathcal T}
\newcommand{\cB}{\mathcal B}
\newcommand{\cP}{\mathcal P}
\newcommand{\cI}{\mathcal I}
\newcommand{\R}{\mathbb R}
\icmltitlerunning{Unified Dynamics in Neural Networks}
\begin{document}
\twocolumn[
  \icmltitle{Unified Dynamics of Training, Learning, and Inference in Neural Networks}
  \begin{icmlauthorlist}
    \icmlauthor{Mian Wang}{whpu}
  \end{icmlauthorlist}
  \icmlaffiliation{whpu}{Wuhan Polytechnic University, Wuhan, China}
  \icmlcorrespondingauthor{Mian Wang}{wangmian945@gmail.com}
  \icmlkeywords{training, learning, inference, feedback, reinforcement learning, unified dynamics}
  \vskip 0.3in
]
\printAffiliationsAndNotice{}
\begin{abstract}
Despite the remarkable success of neural networks, their internal mechanisms remain largely a black box. What determines the outcomes of training? Why can subsequent training of the same model, with identical parameters and training data, still yield different results? What internal changes occur in a model during learning? What is a neural network's capability, and why can capabilities be acquired, lost, or recovered during training? How are inference and learning related? To address these questions, we conducted systematic causal experiments across six neural-network systems, progressively establishing the fundamental relations among training, learning, and inference (TLI) and validating them across architectures. We found that the same realized training update can produce different functional responses in different receiving states and exhibit state-dependent nonlinear behavior; during learning, distributed functional support within neural networks undergoes reorganization; capability is a target-specific decision on a continuous functional readout, with its acquisition, loss, and recovery determined by functional responses crossing decision boundaries; and inference recruits and combines functional support formed through learning while preserving the learned state. Based on the established TLI relations, we developed a prediction algorithm that uses pre-update information and an already formed training update to predict post-update target correctness and capability transitions without accessing actual post-update target outputs. The algorithm achieved prediction accuracies ranging from 91.43\% to 99.87\% across six neural-network systems. Building on these findings, we further found in a seventh, reinforcement-learning system that even when feedback is consistently correct, concentrating it on one capability shifts internal functional support toward the reinforced capability and weakens other previously acquired capabilities; rebalancing feedback redistributes that support and restores those capabilities. On the basis of these experiments, we established an abstract, unified system of coupled dynamical equations describing these relations. This experimentally established system of abstract dynamical equations exhibits broad mathematical unity. Substituting the mathematical models and conditions used in the respective original studies into the same equations recovers relationships across ten established theories: Edge of Stability, catastrophic forgetting, task arithmetic, Neural Tangent Kernel theory, recursive model collapse, Gradient Noise Scale, Spectral Bias, maximal update parameterization, dynamical isometry, and double descent. The same substitutions also yield finite-amplitude, receiving-state, and capability-boundary consequences beyond the recovered relations. We also found that the core relationships in these equations can be explained by the most elementary computations of neural networks. These findings advance our understanding of neural networks from black-box behavior toward unified internal mechanisms.

\end{abstract}
\section{Introduction}\label{sec:introduction}

Training, generalization, capability formation, inference, and related topics are central problems in neural-network research. Different fields have developed influential mathematical theories to study these problems, employing distinct mathematical objects to characterize network states, functional changes, and observable outcomes.

In optimization theory, the Neural Tangent Kernel (NTK) describes parameter-induced output changes through the network Jacobian and the corresponding kernel in specified training regimes \cite{jacot2018}. Edge of Stability studies training behavior in relation to the largest Hessian eigenvalue, learning rate, and loss trajectory \cite{cohen2021,cohen2022adaptive,damian2023}. Gradient Noise Scale characterizes the covariance of stochastic gradients relative to their mean and its dependence on minibatch size \cite{mccandlish2018}. Here, the current state is represented by parameters, optimizer memory, and local derivatives or curvature; realized updates are produced by the optimizer; and responses appear as changes in model outputs or loss.

In representation learning and scaling, Spectral Bias examines frequency-dependent learning rates of functions or kernel modes \cite{rahaman2019}. Dynamical Isometry studies the propagation of signals and gradients through the singular-value structure of layered Jacobians \cite{pennington2017}. Maximal Update Parameterization ($\mu$P) specifies width scalings that preserve nondegenerate feature and output changes as networks grow \cite{yang2021}. Their internal organizations take the form of spectral modes, layerwise Jacobians, and width-scaled feature dynamics; their functional changes appear in learning speed, propagated responses, and predictive performance.

In generalization and model combination, Double Descent studies how test error can rise near interpolation and fall again as model size, training time, or data conditions change \cite{belkin2019,nakkiran2020}. Task Arithmetic constructs parameter displacements from a common pretrained state and combines them to edit several task behaviors \cite{ilharco2023task}; tangent-space studies relate successful combinations to function-space localization \cite{ortiz2023task}. These studies characterize the current state through the interpolation regime or shared pretrained parameters, internal change through solution paths and task vectors, and observable capability through target-specific accuracy and generalization error.

In continual learning, catastrophic forgetting occurs when training on a new task reduces performance on earlier tasks. Elastic Weight Consolidation (EWC) estimates parameter importance using Fisher information from earlier tasks and penalizes changes to important parameters \cite{kirkpatrick2017}. Gradient Episodic Memory (GEM) constrains new-task gradients using their inner products with previous-task gradients \cite{lopez2017}. Orthogonal Gradient Descent (OGD) projects new gradients into directions that approximately preserve earlier-task outputs \cite{farajtabar2020}. In these methods, the existing state includes previously learned parameters and function-sensitive directions; updates are constrained new-task actions; interactions occur through overlapping gradient or functional subspaces; and capability changes are measured as retention, forgetting, or transfer.

In inference research, Attention uses queries, keys, and values to compute input-dependent information weights \cite{vaswani2017}. In-context learning studies how examples and task information provided in the context affect predictions while model parameters remain fixed \cite{brown2020}. Chain-of-Thought prompting studies how intermediate reasoning steps affect performance on complex tasks \cite{wei2022}. Here, the trained model is the fixed state; the query and context condition inference; attention, activations, and intermediate steps constitute transient computations; and functional outcomes are observed in the resulting predictions.

In reinforcement and consequence-driven training, reinforcement learning from human feedback (RLHF) uses preference and reward signals to adjust model behavior \cite{ouyang2022}. Self-training creates further training data from a model's predictions \cite{xie2020}. Recursive model-collapse studies examine the loss of low-probability modes when successive models are trained on data generated by their predecessors \cite{shumailov2024,fu2024}. These settings use feedback from rewards, preferences, or generated-data distributions.

These studies use different mathematical objects---kernels, Hessians, gradients, layer Jacobians, task vectors, task scores, attention weights, and reward signals---yet six recurring categories can be identified: current state, realized update, functional response, internal organization, observable capability, and feedback. \emph{Therefore, a unified description of neural networks is possible. The core relations among Training, Learning, and Inference (TLI) were first established through causal experiments; subsequent cross-field comparison revealed their connections to existing theories.}

The experimental work began in nanoGPT, where new measurement questions emerged during the investigations and prompted the development of new experimental instruments. Through eight experiments involving realized-update interventions, nonlinear response measurements, reversible component interventions, and frozen-state inference tests, we uncovered internal dynamical phenomena and progressively established the fundamental relations among training, learning, and inference. These findings also led to an algorithm for predicting immediate post-update capability transitions. We subsequently tested the established relations and their predictions across five additional neural-network architectures. Building on the TLI findings, we further derived a prediction about behavior-consequence feedback and tested it through controlled reinforcement-learning experiments in a seventh system, revealing how feedback allocation affects existing functional support and changes in capability. These experimental discoveries exposed the limitations of existing descriptions and necessitated a new set of scientific coordinates for investigating neural networks. On this basis, we established a unified system of coupled dynamical equations. Starting from these equations, we recovered relationships across ten established theories under their respective mathematical models and conditions, derived further mathematical results, and ultimately traced the core relations back to elementary neural-network computations.

\section{Experimental Methods and Findings}\label{sec:experiments}
\begin{figure}[!tbp]
\centering
\includegraphics[width=\columnwidth]{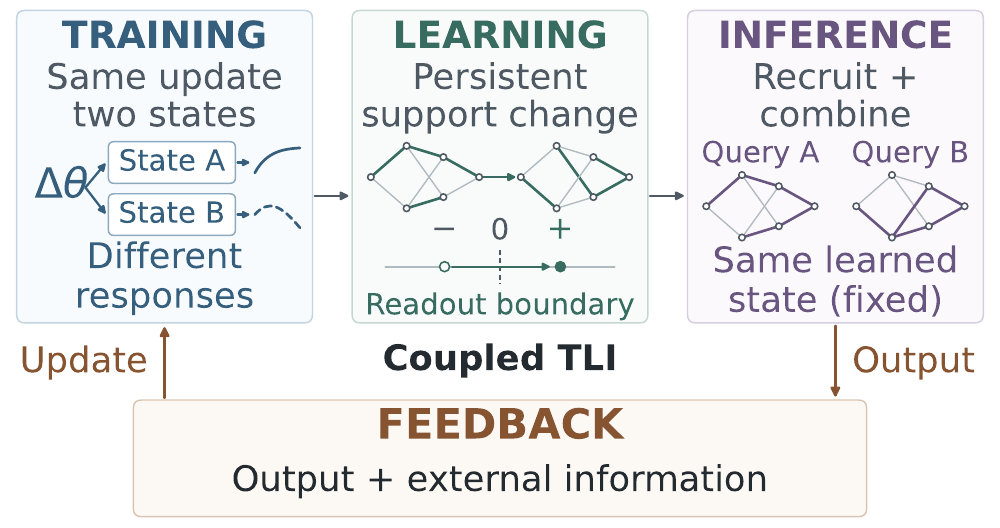}
\caption{Training--Learning--Inference (TLI) with feedback. Realized training updates produce receiving-state-dependent nonlinear functional responses; during learning, distributed functional support undergoes persistent reorganization, while capability acquisition, loss, and recovery depend on whether functional responses cross decision boundaries; inference recruits and combines existing functional support while keeping the learned state unchanged. Model outputs and external information inform subsequent training.}\label{fig:tli-core-feedback}
\end{figure}
\subsection{Experimental instruments and evidence control}
This study uses the Generation-Fact Graph (GFG) as a common experimental record and evidence-linking framework, connecting model and optimizer states, realized updates, interventions, outputs, and their source identities. Four instruments were used to measure key relations in training, learning, and inference.

The realized-update causal-fork instrument restores matched receiving states and compares full and skipped updates together with parameter and optimizer-state exchanges, separating the effects of realized updates from those of the states receiving them. The finite-amplitude update-path instrument applies the same realized update at different amplitudes to a fixed receiving state, recording functional-response trajectories and capability transitions. CSRG-4C uses baseline, single-component, and paired-component gating interventions to measure necessity, backup, substitution, synergy, and failure slack in functional support. The identity-aligned target-boundary ledger aligns target identities and decision margins before and after an update to distinguish capability maintenance, loss, non-acquisition, and acquisition or recovery. All experiments were checked using frozen protocols, control conditions, and independent validators.

\subsection{Experimental establishment of core TLI relations}
The eight nanoGPT experiments that established TLI involved a substantially broader set of native interventions, controlled comparisons, and independent validations than can be presented in the main text. INF-E01 alone examined 52 checkpoints across 13 training histories, executing 520 single- and paired-component gating interventions and 52 component-version rollback and restoration comparisons, with repeated baseline forwards, parameter-version identity checks, and independent validation. This section summarizes their key experimental findings and theoretical relations. Complete protocols, execution records, negative results, and independent validators are provided in the accompanying evidence repository.

\subsubsection{Training: realized updates and receiving states}
At a registered training step, the parameter--optimizer state is $Z_t=(\theta_t,\omega_t)$ and the realized parameter update is $\Delta\theta_t=\theta_{t+1}-\theta_t$. TL-E01 revealed capability formation, decline, and recovery that could not be fully characterized by loss, accuracy, or training step alone. TL-E02 showed that pausing effective parameter--optimizer evolution delays capability formation, while the clipping-threshold control did not reproduce this effect.

TL-E03 used full/skip branches and reciprocal update transplants to establish that the same realized update produces different functional responses in different receiving states:
\begin{equation}\label{eq:exp1}
\begin{aligned}
R_q(\theta_A,\Delta\theta_B)&\quad\text{versus}\quad R_q(\theta_B,\Delta\theta_B),\\
R_q(\theta,\Delta)&=m_q(\theta+\Delta)-m_q(\theta).
\end{aligned}
\end{equation}
Separate optimizer-memory exchanges examined how historical optimizer states affect update formation:
\begin{equation}\label{eq:exp2}\Delta^{(a,b)}=\OptStep(\theta_t,g_t,\omega_{a,b}).\end{equation}
TL-E04 measured the full finite-amplitude response to an actual update:
\begin{equation}\label{eq:exp3}r_{t,q}(\alpha)=m_q(\theta_t+\alpha\Delta\theta_t)-m_q(\theta_t).\end{equation}
The experiments identified near-linear, saturating, accelerating, turn-back, and sign-reversal responses.

\subsubsection{Internal changes during learning}
TL-E05 used reversible component interventions to establish distributed functional support, including component necessity, backup, substitution, synergy, and failure slack. TL-E06 then examined how realized training updates persistently reorganize this support:
\begin{equation}\label{eq:exp4}\Delta P_{t,q}=P_q(\theta_t+\Delta\theta_t)-P_q(\theta_t).\end{equation}
Across 72 experimental sections, 1,636 valid support-allocation transitions and 228 primary-support hand-offs were recorded. TL-E07 linked these internal changes to observable capability through identity-aligned target readout boundaries. The signs of pre- and post-update margins distinguish maintenance, loss, non-acquisition, and acquisition or recovery. The experiments reveal that during learning, distributed functional support within neural networks undergoes persistent reorganization, while capability transitions are determined by target-specific readout-boundary crossings.

\subsubsection{Inference: frozen recruitment of learned support}
INF-E01 examined 13 nanoGPT training histories and 52 checkpoints. With persistent learned states frozen, the experiment measured query-conditioned component effects, non-additive support interactions, and changes produced by replacing trained components with their exact pre-formation versions. At a fixed learned state $L^*$, inference recruits and combines learned support:
\begin{equation}\label{eq:exp5}S_{q,c}^{\mathrm{active}}=\Pi_{q,c}(L^*),\qquad y_{q,c}=H_{q,c}(S_{q,c}^{\mathrm{active}}).\end{equation}
Non-additive component interactions are measured by
\begin{equation}\label{eq:exp6}I_{ij,q}=f_{\mathrm{full}}-f_{-i}-f_{-j}+f_{-i,-j},\end{equation}
while learned-version dependence is tested through
\begin{equation}\label{eq:exp7}D_{j,q}=f(q;\theta^*)-f(q;\theta^{(j\leftarrow\mathrm{pre})}).\end{equation}
Exact restoration verified that these interventions did not persistently alter the learned state. The experiments establish inference as the query-conditioned recruitment and combination of support formed through learning. Together, these eight experiments established the core Training--Learning--Inference (TLI) relations.

\subsection{Prediction and cross-system validation}
\subsubsection{Theory-derived target-boundary prediction}
Based on the established relations among realized training updates, functional responses, and capability readouts, we developed TL-P01 to predict immediate post-update capability transitions. For target $y$, competitor $c$, hidden representation $h$, let $b_c=w_y-w_c$ and $\Delta b_c=\Delta w_y-\Delta w_c$. Using first and second directional derivatives $h'=Dh[\Delta\theta_t]$ and $h''=D^2h[\Delta\theta_t,\Delta\theta_t]$, the predictor computes
\begingroup
\renewcommand{\minalignsep}{0pt}
\begin{flalign}\label{eq:exp8}
& \widehat g_{y,c}=b_c^\top h+\Delta b_c^\top h +b_c^\top h' +\Delta b_c^\top h'+\tfrac12 b_c^\top h''. &&
\end{flalign}
\endgroup
The target is predicted correct when $\min_{c\ne y}\widehat g_{y,c}>0$. Predicted post-update correctness, combined with the known pre-update state, gives four capability transitions. Predictions use the already formed update and pre-response information without reading actual post-update target outputs.

\subsubsection{Cross-system validation}
We tested the established TLI relations and predictions in five additional systems. Appendix Tables~\ref{tab:training-state}--\ref{tab:pred} report the six TLI systems and the separate GRU feedback experiment. Each additional TLI system was examined through realized-update interventions, receiving-state comparisons, finite-amplitude responses, component-coalition support measurements, target-boundary outcomes, and frozen-inference tests. All 16 coalitions of four registered components were evaluated, with exact Shapley effects:
\begin{equation}\label{eq:exp9}
\phi_{j,q}=\sum_{C\subseteq\mathcal C\setminus\{j\}}
\tfrac{|C|!(3-|C|)!}{4!}
\bigl[v_q(C\cup\{j\})-v_q(C)\bigr].
\end{equation}
Here $v_q(C)$ is the target margin under active component coalition $C$, with $|\mathcal C|=4$.

All 42 ResNet, 18 diffusion U-Net, and 36 Vision-MoE parameter receiving-state exchanges exceeded their registered response-difference thresholds. Finite-amplitude nonlinear responses occurred in 32.23\%, 19.25\%, and 35.71\% of their respective target-update records, while support reallocation exceeded its specified threshold in 91.41\%, 97.02\%, and 80.43\%. Frozen-inference experiments further confirmed query-conditioned support, non-additive interactions, and exact learned-component version dependence.

Three GNN seeds met the same eight registered criteria. The remaining criterion, at least 10\% nonlinear responses, was met in one seed (36/256); the other two gave 6/256 and 20/256. The hard-spiking SNN met all twelve registered TLI criteria in all four seeds. The prediction algorithm was also evaluated across these systems with task-specific adaptations of the directional response computation. Table~\ref{tab:pred} reports immediate post-update target-boundary classification. The predictor achieved high four-way recall in five systems. In the hard-spiking SNN, local derivative-based prediction achieved 93.24\% accuracy and 60.05\% four-way recall. Figure~\ref{fig:cross} summarizes the six within-system TLI experiment chains.

\subsection{Effects of Feedback Allocation on Functional Support and Existing Capabilities}
\subsubsection{Theory-derived prediction of feedback dynamics}
The established TLI experiments reveal persistent reorganization of distributed functional support during learning, while inference recruits and combines this learned support to produce behavior. When the consequences of inference are repeatedly fed back into subsequent training, the resulting updates can further reorganize the support on which future inference depends. This led to a prediction: sustained, narrowly concentrated correct positive feedback could progressively favor the reinforced capability while weakening other already mastered capabilities. The predicted effect arises from feedback-directed functional reorganization, rather than incorrect rewards.

\subsubsection{Experimental tests and results}
RL-E05/E06 tested this prediction in a seventh system, a shared attention-free GRU policy. All four skills reached mastery before the experimental branches began. Every branch started from an identical parameter--AdamW state, and each complete two-decision episode received $r=1$ only when correct and $r=0$ otherwise. RL-E05 compared selective, balanced, and frozen feedback. In RL-E06, each of 3,200 updates allocated 60 episodes among four skills $(q_0,q_1,q_2,q_3)$: balanced $(15,15,15,15)$, mildly concentrated $(30,10,10,10)$, highly concentrated $(45,5,5,5)$, and exclusive $(60,0,0,0)$.

RL-E05 observed relative support concentration, cross-skill interference, and component-version-dependent trade-offs. Its stronger preregistered temporal-precedence criterion was not met. In RL-E06, all 12 formal seeds exhibited dose-ordered changes. The reinforced skill's mean support share increased from 25.54\% under balanced feedback to 35.20\% under exclusive feedback, while the other skills' mean margins decreased from 9.474 to 3.962. Mild and high concentration already reduced these margins without reducing accuracy. Exclusive feedback decreased other-skill accuracy by 39.06 percentage points relative to balanced feedback. Two recovery branches started from the exact exclusive-feedback state at update 800. Rebalanced feedback restored other-skill accuracy from 70.31\% to 99.48\%, preserving the reinforced skill at 100\%. Training only the other three skills restored their accuracy to 99.83\% but reduced the previously reinforced skill to 89.06\%.

These experiments establish that, in the tested shared system, the distribution of correct feedback can redirect persistent functional support, alter previously mastered capabilities, and reverse these changes through feedback redistribution.

\begin{figure*}[t]
\begin{minipage}{\textwidth}
\centering
\makebox[\textwidth][c]{\includegraphics[width=\textwidth]{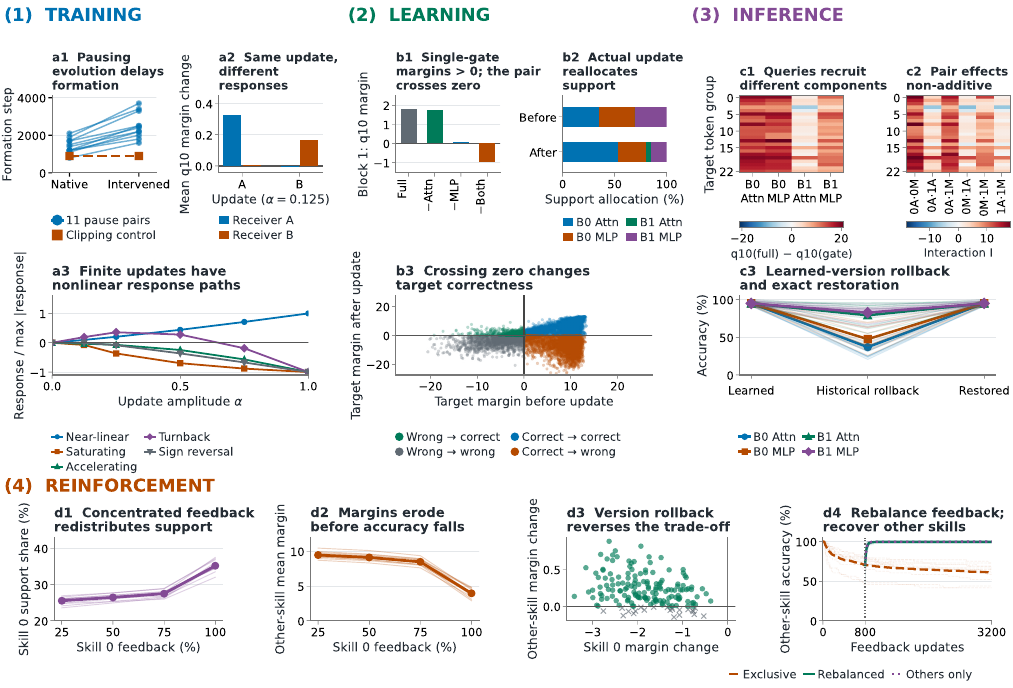}}
\caption{Experimental evidence for Training, Learning, and Inference (TLI), together with reinforcement-learning feedback effects. (a1--a3) Training: delayed capability formation, receiver-dependent update responses, and finite-amplitude responses. (b1--b3) Learning: component dependence, support reallocation, and target-boundary transitions. (c1--c3) Inference: query-conditioned support recruitment, non-additive interactions, and learned-version rollback. (d1--d4) Reinforcement-learning experiments: feedback allocation, support redistribution, erosion of other capabilities, and recovery under rebalanced feedback.}
\label{fig:evidence}
\end{minipage}
\end{figure*}



\section{Unified dynamical equations}\label{sec:equations}
The experimentally identified relations can be written as one coupled dynamical system.

\subsection{Objects and readouts}
Let $L_t$ denote the persistent learned state, $\omega_t$ the optimizer memory, $Z_t=(L_t,\omega_t)$ the complete receiving state, $B_t$ the training information supplied as feedback, and $A_t$ the realized training action. In a parameterized implementation, $\theta_t$ denotes parameters and $\delta_t=\theta_{t+1}-\theta_t$ is the realized parameter increment. For a registered target $q$, $m_q(L)$ is its continuous functional readout and $\chi_q$ the observable capability rule. A fixed intervention protocol $\mathcal J_q$ measures distributed functional support $\mathcal P_q(L;\mathcal J_q)$. For a query and context $(x,c)$, $\Pi_{x,c}$ recruits active support and $H_{x,c}$ combines it; $E_t$ denotes external reference information or environmental consequences used to assess the output.

For classification a concrete signed margin is
\begin{equation}\label{eq:margin}
\begin{aligned}
m_q(L)&=z_y(L,x)-\max_{k\ne y}z_k(L,x),\\
\chi_q(m)&=\mathbf 1[m>0],\quad q=(x,y).
\end{aligned}
\end{equation}
The registered readout in other systems may be different; the diffusion experiment uses a frozen residual-error boundary.

\subsection{Why this form is necessary}
The equations retain only the distinctions forced by the experimental counterexamples. Parameter-only gradient dynamics cannot represent optimizer-memory-dependent actions; therefore, the receiving state is $Z_t=(L_t,\omega_t)$, and the realized action is $A_t=\cU(Z_t,B_t)$. Local Hessian and fixed-kernel descriptions cannot determine finite responses under receiving-state dependence, turnback, and sign reversal; therefore, the exact response $R_{t,q}=m_q(L_{t+1})-m_q(L_t)$ is retained. Function-space behavior and fixed-circuit localization cannot represent support reallocation and primary handoff when capability remains unchanged; therefore, learning is measured by $P_{t,q}=\cP_q(L_t;\mathcal J_q)$ and $\Delta P_{t,q}=P_{t+1,q}-P_{t,q}$. Discrete emergent-capability descriptions cannot distinguish continuous readout evolution from a boundary crossing; therefore, capability is written as $C_{t,q}=\chi_q(m_q(L_t))$. Static or additive inference descriptions cannot represent query-conditioned recruitment, non-additive interaction, and learned-version dependence at a fixed learned state; therefore, inference requires $S_{x,c}^{\mathrm{active}}=\Pi_{x,c}(L^*)$, $Y_{x,c}=H_{x,c}(S_{x,c}^{\mathrm{active}})$, and $\Delta_{\mathrm{inf}}L=0$. Scalar reward correctness cannot represent allocation-dependent capability erosion and recovery under entirely correct feedback; therefore, $B_{t+x}=\cB(Y_t,E_t)$ must explicitly determine later training actions and state transitions.

\subsection{Training: realized action and persistent state}
\begin{equation}\label{eq:training}
\begin{aligned}
Z_t&=(L_t,\omega_t),\\
A_t&=\cU(Z_t,B_t),\\
Z_{t+1}&=\cT(Z_t,A_t).
\end{aligned}
\end{equation}
The receiving state matters both for action formation and for the action's realized effect. Optimizer-memory exchanges in the experimental program separate the stored optimizer history from the receiving parameters. For parameterized networks the same transition is represented as
\begin{equation}\label{eq:update}
\delta_t=\Psi(\theta_t,\omega_t,B_t),\qquad
\theta_{t+1}=\theta_t+\delta_t.
\end{equation}
Here $B_t$ includes training information and its evaluative signals, such as supervised input--target comparisons or reinforcement rewards. $\Psi$ names the executed optimizer-specific update formation; the functional consequence of that update is a separate observable.

\subsection{Functional response, internal changes during learning, and capability}
\begin{equation}\label{eq:response}
R_{t,q}=m_q(L_{t+1})-m_q(L_t).
\end{equation}
\begingroup
\renewcommand{\minalignsep}{0pt}
\begin{flalign}\label{eq:support}
& P_{t,q}=\cP_q(L_t;\mathcal J_q),\qquad \Delta P_{t,q}=P_{t+1,q}-P_{t,q}. &&
\end{flalign}
\endgroup
\begin{equation}\label{eq:capability}
\begin{aligned}
C_{t,q}&=\chi_q\bigl(m_q(L_t)\bigr),\\
C_{t+1,q}&=\chi_q\bigl(m_q(L_t)+R_{t,q}\bigr).
\end{aligned}
\end{equation}
The response $R_{t,q}$, persistent support change $\Delta P_{t,q}$, and capability readout $C_{t,q}$ are distinct measurements of the same realized state transition. The experiments reveal that during learning, distributed functional support within neural networks undergoes persistent reorganization. Changes in internal support can occur with a stable discrete correctness readout.

\subsection{Inference: recruitment from a fixed learned state}
\begin{equation}\label{eq:inference}
\begin{aligned}
S_{x,c}^{\mathrm{active}}&=\Pi_{x,c}(L^*),\\
Y_{x,c}&=H_{x,c}\bigl(S_{x,c}^{\mathrm{active}}\bigr),\\
\Delta_{\mathrm{inf}}L&=0.
\end{aligned}
\end{equation}
Here $L^*$ is fixed during a registered inference. $\Pi_{x,c}$ denotes query-conditioned selection or recruitment, including nonlinear operations; it is not restricted to a linear projector. Recurrent computation, token generation, denoising steps, and routing may change transient activations while the persistent learned state remains fixed.

\subsection{Feedback and subsequent training}
\begin{equation}\label{eq:feedback}
\begin{aligned}
Y_t&=\cI(L_t,x_t,c_t),\\
B_{t+x}&=\cB(Y_t,E_t),\\
A_{t+x}&=\cU(Z_{t+x},B_{t+x}),\\
Z_{t+x+1}&=\cT(Z_{t+x},A_{t+x}).
\end{aligned}
\end{equation}
Here $x\geq1$ denotes the feedback delay in the chosen step indexing, distinct from the model input $x_t$. Under the supervised-learning convention, $x=1$; in reinforcement learning, the delay is determined by the feedback-return mechanism and can vary between interactions. The feedback $B_{t+x}$ is formed from the model output $Y_t$ and external information $E_t$, without requiring prior feedback as an input. In supervised training, $\cB$ can compare model outputs with supplied targets; in reinforcement learning, it can incorporate environmental consequences such as rewards. In both cases the resulting feedback informs subsequent training actions, whose updates change later learned states and future inference. The experimental anchors are TL-E01--04 for training and response; TL-E05--07 for support and capability; INF-E01 for frozen inference; and RL-E05/06 for consequence-mediated feedback closure. Together, equations~\eqref{eq:training}--\eqref{eq:feedback} comprise the unified system of dynamical equations.

\section{Mathematical unification}\label{sec:unification}
Starting from the experimentally established unified TLI dynamical equations, we combine them with the mathematical models and conditions used in the respective original studies to derive mathematical relationships across ten established theories. One complete derivation is presented below; the remaining nine appear in Appendix~\ref{app:all10}. The markers \Tmark, \Amark, and \Rmark{} denote TLI equations, source-model conditions, and recovered relations. The marker \Rplusmark{} denotes an additional consequence obtained by combining a TLI relation with the stated source-model conditions.

\subsection{Edge of Stability: optimization and capability stability}\label{sec:ex:eos}
\textbf{TLI equations \Tmark.} Equations~\eqref{eq:training}--\eqref{eq:capability} couple the complete receiving state to the realized action and capability:
\begin{sourceequation}{T}\label{eq:eos-T}
\begin{aligned}
Z_t&=(L_t,\omega_t),\quad \theta_{t+1}=\theta_t+\delta_t,\\
C_{t+1,q}&=\chi_q(m_q(L_t)+R_{t,q}).
\end{aligned}
\end{sourceequation}
\textbf{Source-model conditions \Amark.} In a positive eigenmode of the local fixed-quadratic model \cite{cohen2021,damian2023}, let the error be $e_t$ and curvature $\lambda>0$. Gradient descent gives $e_{t+1}=(1-\eta\lambda)e_t$. Classical heavy-ball momentum \cite{polyak1964} gives $v_{t+1}=\beta v_t-\eta\lambda e_t$ and $e_{t+1}=e_t+v_{t+1}$, with $0\leq\beta<1$.

\textbf{Substitution and recovery \Rmark.} The gradient-descent mode contracts under the Edge-of-Stability condition
\begin{sourceequation}{R}\label{eq:eos-boundary}
|1-\eta\lambda|<1\quad\Longleftrightarrow\quad0<\eta\lambda<2,
\end{sourceequation}
and the heavy-ball state is asymptotically stable for $0<\eta\lambda<2(1+\beta)$.

\textbf{Capability consequence \Rplusmark.} For $m_q=e$ and $\chi_q(m)=\mathbf1[m>0]$,
\begin{sourceequation}{R+}\label{eq:eos-memory-boundary}
\begin{aligned}
C_{t,q}^{\rm GD}&=\mathbf1[(1-\eta\lambda)^t e_0>0],\\
C_{t+1,q}^{\rm HB}&=\mathbf1[(1-\eta\lambda)e_t+\beta v_t>0].
\end{aligned}
\end{sourceequation}
Thus $1<\eta\lambda<2$ contracts the gradient-descent error while the capability alternates. At $e_t=0.1$, $\eta\lambda=0.5$, and $\beta=0.9$, heavy-ball memories $v_t=0.1$ and $v_t=-0.1$ produce margins $0.14$ and $-0.04$ inside the stable region. Optimization stability and capability preservation have different boundaries, and the latter depends on optimizer memory. Experimental anchors: TL-E01--04 and TL-E06--07.

\section{This Is a Simple Theory}\label{sec:simple}
The final finding, reached after causal experiments, cross-architecture tests, feedback validation, and theory-first extrapolations, is that the central relations can be recovered from elementary neural-network computation and parameter updates.

\subsection{One update can produce opposing functional responses}
Consider three targets sharing two parameters:
\begin{equation}\label{eq:simple-targets}
f_A=a+b,\qquad f_B=a,\qquad f_C=1-b.
\end{equation}
Starting at $(a,b)=(0.2,0.1)$, a gradient-descent step on $\ell_A=\frac12(f_A-1.3)^2$ with $\eta=0.2$ gives $(a,b)\mapsto(0.4,0.3)$. Thus
\begin{equation}\label{eq:simple-opposing}
(\Delta f_A,\Delta f_B,\Delta f_C)=(+0.4,+0.2,-0.2).
\end{equation}
A single update formed from target $A$ improves one untargeted score while reducing another because the parameters participate in shared computations.

\subsection{Nonlinear responses depend on state and update}
For $f_q(a,b)=ab$, apply the actual increment $(\delta a,\delta b)$ at receiving state $(a,b)$. Scaling the increment by $\alpha$ gives the exact response
\begin{equation}\label{eq:simple-nonlinear}
R_q(\alpha)=\alpha(b\delta a+a\delta b)+\alpha^2\delta a\,\delta b.
\end{equation}
The coefficients depend on the receiving state; the cross-term makes the finite-amplitude response nonlinear whenever $\delta a\,\delta b\ne0$.

\subsection{Parameter updates redistribute functional support}
Return to $f_A=a+b$. Under the fixed intervention protocol $\mathcal J_A$ that removes each additive component's contribution in turn, the measured contributions change from $(0.2,0.1)$ to $(0.4,0.3)$. Their normalized shares change from $(2/3,1/3)$ to $(4/7,3/7)$. Thus $\Delta P_A\ne0$ under the same intervention protocol, even though both components remain present. Shared computation and ordinary parameter updates provide an elementary mechanism for persistent redistribution of functional support.

\subsection{Capability changes arise from readout-boundary crossings}
Define the signed margin for target $B$ as $m_B=f_B-0.3$ and correctness as $C_B=\mathbf1[m_B>0]$. The same update moves $m_B$ from $-0.1$ to $+0.1$, hence $C_B$ changes from $0$ to $1$. In general,
\begingroup
\thickmuskip=2mu
\medmuskip=2mu
\renewcommand{\minalignsep}{0pt}
\begin{flalign}\label{eq:simple-crossing}
& m_q(L_t)\le0<m_q(L_t)+R_{t,q}
\Rightarrow C_{t,q}=0,\, C_{t+1,q}=1. &&
\end{flalign}
\endgroup

The discrete capability transition reads out the continuous response; it is not a separate stored state variable.

\subsection{Inference recruits and combines learned computation}

Consider a single nonlinear neuron with two learned input weights:
\begin{equation}\label{eq:simple-relu}
f_\theta(x)=\operatorname{ReLU}(ax_1+bx_2).
\end{equation}

At the fixed learned state $\theta^*=(2,1)$, the two component contributions are $s_1(x)=2x_1$ and $s_2(x)=x_2$, and inference produces
\begin{equation}\label{eq:simple-inference-fixed}
Y_x=\operatorname{ReLU}(s_1(x)+s_2(x)),
\qquad \Delta_{\mathrm{inf}}L=0.
\end{equation}

The same learned components contribute differently under different inputs. Their combination can also be non-additive. To measure this, introduce reversible component gates $g_1,g_2\in\{0,1\}$ and evaluate
\begin{equation}\label{eq:simple-gating}
f_{g_1,g_2}(x)=
\operatorname{ReLU}(2g_1x_1+g_2x_2).
\end{equation}

For $x=(-1,1)$, the four gated outputs $(f_{11},f_{10},f_{01},f_{00})$ are $(0,0,1,0)$, giving the exact interaction
\begin{equation}\label{eq:simple-inference}
I_{12}=f_{11}-f_{10}-f_{01}+f_{00}=-1.
\end{equation}

For $x=(1,1)$, the same interaction is zero. Thus query-dependent functional contributions and non-additive component effects arise from elementary nonlinear computation with unchanged learned parameters.

Inference also depends on the versions formed through learning. Replacing the trained first-component weight $a^*=2$ with a pre-learning version $a_{\mathrm{pre}}=1$ changes the output at $x=(1,1)$ from $3$ to $2$; exact restoration recovers the original result. The gates and version replacements are diagnostic interventions, distinct from ordinary inference.

These calculations illustrate the TLI inference relation: a fixed learned state supplies functional components, the current input determines their contributions, and nonlinear computation combines them into an output. No persistent learning update is required.

\subsection{Feedback and subsequent training}

When model outputs are compared with supervised targets or assessed through environmental consequences, the resulting feedback can change subsequent training and the learned computation available to later inference:
\begin{equation}\label{eq:simple-closure}
(Y_t,E_t)\longrightarrow B_{t+x}
\longrightarrow A_{t+x}\longrightarrow Z_{t+x+1}.
\end{equation}

Inference preserves the current learned state; feedback-driven training can change the learned state used by future inference.

\section{Conclusion}\label{sec:conclusion}
Through systematic causal experiments, we established and validated the fundamental relations among training, learning, and inference (TLI) across six neural-network systems. Realized training updates produce receiving-state-dependent nonlinear functional responses; during learning, distributed functional support undergoes persistent reorganization, while capability acquisition, loss, and recovery depend on whether functional responses cross decision boundaries; inference recruits and combines existing functional support while keeping the learned state unchanged. In a seventh, reinforcement-learning system, we further found that even when feedback is consistently correct, concentrating it on one capability shifts functional support toward the reinforced capability and weakens other previously acquired capabilities, whereas rebalancing feedback restores those capabilities.

We formulated these experimentally established relationships as a unified system of coupled dynamical equations. Starting from these equations and using the mathematical models and conditions of the respective studies, we recovered relationships across ten established theories and derived further consequences for finite-amplitude response, receiving-state dependence, and capability boundaries. The core relationships can also be recovered from elementary neural-network computations. Together, these findings move the study of neural networks from black-box observation toward a mechanistic understanding of internal dynamics and connect previously separate research fields through a unified mathematical framework.

\clearpage
\section*{Impact Statement}
Through causal experiments across six neural-network systems, this work establishes and validates the fundamental relations among training, learning, and inference. Experiments in a seventh reinforcement-learning system further reveal how the allocation of correct feedback affects functional support and previously acquired capabilities. These findings are expressed through a unified system of coupled dynamical equations, connecting previously separate neural-network research fields and providing a mathematical framework for understanding their internal mechanisms.

\section*{AI Disclosure Statement}
General-purpose AI systems, including OpenAI Codex and ChatGPT, served as scientific agents within the GFG-based recursive scientific process. Under the author's direction, these AI systems invented and implemented the four experimental instruments---CSRG-4C, realized-update causal forks, finite-amplitude update paths, and the identity-aligned target-boundary ledger---and designed, implemented, initiated, monitored, and executed the reported experiments. The ten mathematical extrapolations reported in this paper were derived primarily by these AI scientific agents, which also designed and implemented their executable small-network validation suite. They performed analyses, causal interventions, matched replays, counterexample searches, and independent validation procedures. The reported numerical results were produced by the documented native programs and execution environments, not generated directly as text by AI models. The author defined the scientific questions, research boundaries, and decision criteria, reviewed the validated evidence, and accepts responsibility for the scientific claims and results.

\begingroup
\emergencystretch=2em
\setlength{\bibhang}{10pt}
\bibliography{references}
\bibliographystyle{icml2027}
\endgroup
\clearpage
\appendix
\onecolumn
\section{Remaining complete mathematical derivations (9 cases)}\label{app:all10}
The cases presented here retain the source-model conditions and explicit calculations, with markers \Tmark, \Amark, \Rmark, and \Rplusmark. Together with the derivation in Section~\ref{sec:unification}, they constitute the ten mathematical specializations.
\subsection{Catastrophic Forgetting: finite-amplitude reversal of gradient alignment}\label{sec:ex:forgetting}
\textbf{TLI equations \Tmark.} The exact continuous response and target-specific correctness readout are
\begin{sourceequation}{T}\label{eq:forgetting-T}
\begin{aligned}
R_{t,q}&=m_q(L_{t+1})-m_q(L_t),\\
C_{t,q}&=\mathbf1[m_q(L_t)>0].
\end{aligned}
\end{sourceequation}
\textbf{Original mathematical conditions \Amark.} For the gradient-based continual-learning specialization, a previously mastered target has $m_q(L_0)>0$. A later task $B$ supplies the source-model action $\delta_t=-\eta\nabla_\theta L_B(\theta_t)$, with differentiable target margins and small steps for the \emph{local gradient approximation} \cite{kirkpatrick2017,lopez2017,farajtabar2020}.

\textbf{Substitution and recovery \Rmark.} Telescoping the \emph{exact} TLI response along the actually executed sequence yields
\begin{sourceequation}{R}\label{eq:forgetting-telescope}
m_q(L_T)=m_q(L_0)+\sum_{t=0}^{T-1}R_{t,q}.
\end{sourceequation}
With the recorded readout convention in \eqref{eq:forgetting-T}, an initially correct target becomes incorrect at step $T$ precisely when
\begin{sourceequation}{R}\label{eq:forgetting-boundary}
\sum_{t=0}^{T-1}R_{t,q}\le -m_q(L_0).
\end{sourceequation}
The telescoping identity and its threshold are exact. For the source-model SGD action, the small-step differentiable specialization gives
\begin{sourceequation}{R}\label{eq:forgetting-gradient}
R_{t,q}\simeq-\eta\nabla_\theta m_q(\theta_t)^\top\nabla_\theta L_B(\theta_t),
\end{sourceequation}
and, taking an older task's differentiable loss itself as the readout,
\begin{sourceequation}{R}\label{eq:forgetting-oldloss}
\Delta L_{\mathrm{old}}\simeq -\eta g_{\mathrm{old}}^\top g_{\mathrm{new}}.
\end{sourceequation}
Gradient alignment determines the local contribution, while the exact accumulated signed responses decide the eventual readout-boundary crossing. Subsequent positive responses can cross back and recover a lost target.

\textbf{Finite-amplitude consequence \Rplusmark.} For a twice continuously differentiable margin along the realized parameter segment,
\begin{sourceequation}{R+}\label{eq:forgetting-finite}
\begin{aligned}
R_{t,q}&=G_{t,q}+Q_{t,q},\qquad G_{t,q}=\nabla m_q(\theta_t)^\top\delta_t,\\
Q_{t,q}&=\int_0^1(1-s)\,\delta_t^\top
H_{m_q}(\theta_t+s\delta_t)\delta_t\,ds.
\end{aligned}
\end{sourceequation}
If $m_q(L_0)+\sum_tG_{t,q}>0$, the first-order accumulated prediction is reversed exactly when
\begin{sourceequation}{R+}\label{eq:forgetting-reversal}
\sum_tQ_{t,q}\leq-m_q(L_0)-\sum_tG_{t,q}.
\end{sourceequation}
For $m_q(a,b)=ab-0.99$ at $(a,b)=(1,1)$, a unit-step gradient action for $L_B=[(a-1.3)^2+(b-0.75)^2]/2$ gives $\delta=(0.30,-0.25)$. Here $G_q=0.05$, $Q_q=-0.075$, and $R_q=-0.025$: the first-order margin is $0.06$, while the realized margin is $-0.015$. A locally favorable aligned action therefore crosses the capability boundary through its finite-amplitude term. Experimental anchors: TL-E03--04, TL-E06--07, and RL-E06.

\subsection{Task Arithmetic: nonadditive response and capability preservation}\label{sec:ex:task-arithmetic}
\textbf{TLI equations \Tmark.} For a fixed receiving state $\theta$ and finite increment $w$,
\begin{sourceequation}{T}\label{eq:ta-T}
\begin{aligned}
R_q(\theta,w)&=m_q(\theta+w)-m_q(\theta),\\
C_q(\theta+w)&=\chi_q\!\left(m_q(\theta)+R_q(\theta,w)\right).
\end{aligned}
\end{sourceequation}
A weight edit is a persistent state intervention in these equations.

\textbf{Source-model conditions \Amark.} Task arithmetic combines displacements from a shared pretrained state \cite{ilharco2023task}:
\begin{sourceequation}{A}\label{eq:ta-A}
\begin{aligned}
u&=\theta_1-\theta,\qquad v=\theta_2-\theta,\\
\theta_{\rm merge}&=\theta+u+v.
\end{aligned}
\end{sourceequation}
Under a tangent-linear readout, the individual responses superpose, recovering the additive relation \Rmark. Function-space localization supplies the corresponding task-wise disentanglement condition \cite{ortiz2023task}.

\textbf{Finite-amplitude consequence \Rplusmark.} The exact interaction at the same receiving state is
\begin{sourceequation}{R+}\label{eq:ta-interaction}
\begin{aligned}
I_q(\theta;u,v)
&=R_q(\theta,u+v)\\
&\quad-R_q(\theta,u)-R_q(\theta,v),\\
&=R_q(\theta+u,v)-R_q(\theta,v).
\end{aligned}
\end{sourceequation}
With $m_q^{\rm add}=m_q(\theta)+R_q(\theta,u)+R_q(\theta,v)$, the merged margin and capability are
\begin{sourceequation}{R+}\label{eq:ta-capability}
\begin{aligned}
m_q(\theta_{\rm merge})&=m_q^{\rm add}+I_q,\\
C_q(\theta_{\rm merge})&=\mathbf1[m_q^{\rm add}+I_q>0],\\
C_q(\theta_{\rm merge})=0&\ \Longleftrightarrow\ I_q\leq-m_q^{\rm add}.
\end{aligned}
\end{sourceequation}
For $m_q(a,b)=0.5-ab$ at $\theta=(0,0)$, take orthogonal edits $u=(1,0)$ and $v=(0,1)$. Both individual responses are zero, while $I_q=-1$ and the merged margin is $-0.5$. The individual edits preserve the capability and their sum removes it. Parameter-space orthogonality is compatible with a decisive functional interaction. Experimental connections: TL-E03--04 and TL-E05--07.

\subsection{Neural Tangent Kernel: finite-step response and capability boundaries}\label{sec:ex:ntk}
\textbf{TLI equations \Tmark.} For a parameterized learned state, equations~\eqref{eq:update}--\eqref{eq:capability} give
\begin{sourceequation}{T}\label{eq:ntk-T}
\begin{aligned}
\theta_{t+1}&=\theta_t+\delta_t,\\
R_{t,q}&=m_q(\theta_t+\delta_t)-m_q(\theta_t),\\
C_{t+1,q}&=\chi_q(m_q(\theta_t)+R_{t,q}).
\end{aligned}
\end{sourceequation}

\textbf{Source-model conditions \Amark.} Let $f_\theta$ be differentiable, $F_\theta=f_\theta(X)$ its stacked training outputs, and $J_\theta(X)=\partial F_\theta/\partial\theta$. Under Euclidean gradient descent on $\ell(F_\theta)$, the NTK construction \cite{jacot2018} uses
\begin{sourceequation}{A}\label{eq:ntk-A}
g_t=\nabla_F\ell(F_{\theta_t}),\qquad
\delta_t=-\eta J_{\theta_t}(X)^\top g_t,\qquad
K_\theta(x,X)=J_\theta(x)J_\theta(X)^\top.
\end{sourceequation}

\textbf{Tangent recovery \Rmark.} When the receiving Jacobian is constant over the update segment, substitution into the response equation gives
\begin{sourceequation}{R}\label{eq:ntk-recovery}
R_{t,x}\simeq-\eta K_{\theta_t}(x,X)g_t.
\end{sourceequation}
In the gradient-flow limit, $\dot f_t(x)=-K_{\theta_t}(x,X)g_t$. Under the original infinite-width NTK conditions, the kernel remains constant in the corresponding limit \cite{jacot2018}.

\textbf{Finite-step consequence \Rplusmark.} Continuous differentiability along the realized segment gives the exact path integral
\begin{sourceequation}{R+}\label{eq:ntk-finite}
\begin{aligned}
R_{t,x}&=\int_0^1J_{\theta_t+s\delta_t}(x)\delta_t\,ds
=-\eta\overline K_t(x,X)g_t,\\
\overline K_t(x,X)&=\int_0^1J_{\theta_t+s\delta_t}(x)
J_{\theta_t}(X)^\top\,ds.
\end{aligned}
\end{sourceequation}
The finite-step operator uses a path-averaged receiving Jacobian and the fixed Jacobian that formed the gradient. On the training set it is generally asymmetric, so its geometry differs from a positive-semidefinite Gram matrix. Related exact finite-model path kernels are studied in \cite{bell2023exact}.

For a scalar signed margin, let $j_q=\nabla_\theta m_q(\theta_t)^\top$. If its gradient is Lipschitz on the segment with constant $M_q$, then
\begin{sourceequation}{R+}\label{eq:ntk-remainder}
\begin{aligned}
R_{t,q}&=j_q\delta_t+\varepsilon_{t,q},\\
\varepsilon_{t,q}&=\int_0^1[\nabla m_q(\theta_t+s\delta_t)-\nabla m_q(\theta_t)]^\top\delta_t\,ds,\\
|\varepsilon_{t,q}|&\leq\tfrac12M_q\|\delta_t\|^2.
\end{aligned}
\end{sourceequation}
Put $\widehat m_q=m_q(\theta_t)+j_q\delta_t$ and $b_q=M_q\|\delta_t\|^2/2$. The capability equation yields
\begin{sourceequation}{R+}\label{eq:ntk-capability}
\widehat m_q>b_q\Rightarrow C_{t+1,q}=1,\qquad
\widehat m_q\leq-b_q\Rightarrow C_{t+1,q}=0.
\end{sourceequation}
Inside the interval $[-b_q,b_q]$, the finite response controls the boundary crossing.

For $f(a,b)=ab$, $m_q=f-0.1$, $\ell=f^2/2$, $(a,b)=(1,1)$, and $\eta=0.6$, the realized update is $(-0.6,-0.6)$. The tangent-predicted margin is $-0.3$, while the exact margin is $0.06$ with $\varepsilon_{t,q}=0.36$. The tangent response predicts loss and the realized finite step preserves the capability. Experimental connections: TL-E03--04 and TL-E05--07.

\subsection{Recursive model collapse: feedback allocation and capability boundaries}\label{sec:ex:collapse}
\textbf{TLI equations \Tmark.} The fixed-state inference~\eqref{eq:inference} and feedback closure~\eqref{eq:feedback} give the following relations. Here the time-step index counts model generations and $x$ denotes the corresponding feedback lag:
\begin{sourceequation}{T}\label{eq:collapse-T}
\begin{aligned}
S_{x,c}^{\mathrm{active}}&=\Pi_{x,c}(L^*),\\
Y_{x,c}&=H_{x,c}(S_{x,c}^{\mathrm{active}}),\\
\Delta_{\mathrm{inf}}L&=0,\\
Y_t&=\cI(L_t,x_t,c_t),\\
B_{t+x}&=\cB(Y_t,E_t),\\
A_{t+x}&=\cU(Z_{t+x},B_{t+x}),\\
Z_{t+x+1}&=\cT(Z_{t+x},A_{t+x}).
\end{aligned}
\end{sourceequation}

\textbf{Original mathematical conditions \Amark.} Consider the original finite discrete support with fixed-$M$ independent sampling and exact empirical-frequency fitting model used to analyze recursive replacement \cite{shumailov2024,gerstgrasser2024}. If $p_t(i)=q_t$ is the probability of category $i$, the next-generation count obeys
\begin{sourceequation}{A}\label{eq:collapse-binomial}
N_{t+1,i}\mid p_t\sim\operatorname{Binomial}(M,q_t),\qquad q_{t+1}=N_{t+1,i}/M.
\end{sourceequation}
\textbf{Substitution and recovery \Rmark.} Substituting this source-model inference--sampling--fit map into TLI's closed-loop state transition gives
\begin{sourceequation}{R}\label{eq:collapse-moments}
\Pr(q_{t+1}=0\mid q_t)=(1-q_t)^M,\qquad \mathbb E[q_{t+1}\mid q_t]=q_t,\qquad \Var(q_{t+1}\mid q_t)=q_t(1-q_t)/M.
\end{sourceequation}
The support can be lost stochastically while its conditional mean frequency remains $q_t$. In vector form,
\begin{sourceequation}{A}\label{eq:collapse-mult}
p_{t+1}=M^{-1}\operatorname{Multinomial}(M,p_t).
\end{sourceequation}
For fixed $M$, the empirical distributions occupy a finite state space. Under pure recursive replacement, a missing category remains missing and each point mass $e_i$ is absorbing. From every nonabsorbing state, the probability of drawing all $M$ samples from one category is $\sum_i p_t(i)^M>0$. The finite chain is therefore almost surely absorbed at a point mass; coordinate-wise bounded martingales additionally give
\begin{sourceequation}{R}\label{eq:collapse-absorbing}
p_t\xrightarrow[t\to\infty]{\mathrm{a.s.}} e_I,
\qquad\Pr(I=i)=p_0(i).
\end{sourceequation}
For the original mixed-source variant, let a fixed proportion $\alpha\in(0,1]$ of the sampling distribution come from real data $r$ with $r_i>0$. Then $\widetilde q_t=(1-\alpha)q_t+\alpha r_i$ and
\begin{sourceequation}{R}\label{eq:collapse-mix}
\mathbb E[q_{t+1}]=r_i+(1-\alpha)^{t+1}(q_0-r_i),\qquad \Pr(q_{t+1}>0\mid q_t=0)=1-(1-\alpha r_i)^M>0.
\end{sourceequation}
Thus missing categories cease to be absorbing under continued real-data input.

\textbf{Capability and allocation consequence \Rplusmark.} For category $i$, let $m_{t,i}=q_t-\tau_i$, $0<\tau_i<1$, and $C_{t,i}=\mathbf1[m_{t,i}>0]$. In the mixed-source model, draw $M$ samples from $s_{t,i}=(1-\alpha)q_t+\alpha r_i$ and fit their empirical frequencies. Put $j_i=\lfloor M\tau_i\rfloor$ and $F_i(s)=\sum_{k=0}^{j_i}\binom Mk s^k(1-s)^{M-k}$. Then
\begin{sourceequation}{R+}\label{eq:collapse-cap-prob}
\begin{aligned}
R_{t,i}&=N_{t+1,i}/M-q_t,\\
\Pr(C_{t+1,i}=0\mid p_t)&=F_i(s_{t,i}),\\
\mathbb E[R_{t,i}\mid p_t]&=\alpha(r_i-q_t).
\end{aligned}
\end{sourceequation}
For $0<s_{t,i}<1$ and $\alpha>0$, the loss probability responds to feedback allocation as
\begin{sourceequation}{R+}\label{eq:collapse-allocation}
\frac{\partial F_i(s_{t,i})}{\partial r_i}
=-\alpha M\binom{M-1}{j_i}s_{t,i}^{j_i}(1-s_{t,i})^{M-1-j_i}.
\end{sourceequation}
Under a fixed total mixture weight, transferring allocation away from category $i$ strictly increases its marginal loss probability. For $q_{t,A}=q_{t,B}=0.1$, $\tau_A=\tau_B=0.05$, $M=100$, $\alpha=0.5$, and $r_A+r_B=0.2$, allocation $(0.1,0.1)$ gives approximately $5.76\%$ loss probability for each category. Allocation $(0,0.2)$ changes the two probabilities to approximately $61.60\%$ and $0.16\%$.

For pure replacement ($\alpha=0$), let $T=\inf\{t\geq0:q_t\leq\tau_i\}$ with $q_0>\tau_i$. Optional stopping of the bounded martingale at the first threshold crossing or absorption at one gives
\begin{sourceequation}{R+}\label{eq:collapse-hitting-bound}
1-q_0\leq\Pr(T<\infty)\leq\frac{1-q_0}{1-\tau_i}.
\end{sourceequation}
The lower bound is the eventual extinction probability; the upper bound includes capability loss before extinction. These relations connect recursive sampling, feedback allocation, and the boundary crossings observed in RL-E06.

\subsection{Spectral Bias: modal decay, forcing, and relearning}\label{sec:ex:spectral}
\textbf{TLI equations \Tmark.} The realized update and target response from equations~\eqref{eq:update} and \eqref{eq:response} are:
\begin{sourceequation}{T}\label{eq:spectral-T}
\delta_t=\Psi(\theta_t,\omega_t,B_t),\qquad \theta_{t+1}=\theta_t+\delta_t,\qquad R_{t,q}=m_q(L_{t+1})-m_q(L_t).
\end{sourceequation}

\textbf{Original mathematical conditions \Amark.} Use squared loss, gradient descent, differentiable outputs, and the locally fixed tangent-kernel approximation of spectral learning studies \cite{jacot2018,rahaman2019}. Let $r_t=f_{\theta_t}-y$, $J=\partial f/\partial\theta$, and $K=JJ^\top$ fixed over the analyzed regime. The source-model update and linearized output effect are
\begin{sourceequation}{A}\label{eq:spectral-update}
\delta_t=-\eta J^\top r_t,\qquad
\Delta f_t\simeq J\delta_t=-\eta Kr_t.
\end{sourceequation}
\textbf{Substitution and recovery \Rmark.} The TLI response gives $r_{t+1}\simeq(I-\eta K)r_t$. In the eigendirection $Kv_k=\lambda_kv_k$,
\begin{sourceequation}{R}\label{eq:spectral-decay}
r_{k,t}\simeq(1-\eta\lambda_k)^t r_{k,0}.
\end{sourceequation}
For $0<\eta\lambda_k<1$, larger eigenvalues imply faster modal error decay. The source model identifies the corresponding kernel modes with frequencies.

\textbf{Finite-step consequence \Rplusmark.} Let $K_t=J_tJ_t^\top$ evolve and use the exact squared-loss step $\delta_t=-\eta J_t^\top r_t$. For a fixed eigenbasis $K_0v_k=\lambda_kv_k$, define $a_{k,t}=v_k^\top r_t$. Equation~\eqref{eq:ntk-finite} gives $\Delta f_t=-\eta K_tr_t+\varepsilon_t$, where $\varepsilon_t=f_{\theta_t+\delta_t}-f_{\theta_t}-J_t\delta_t$. Projection onto the reference mode yields
\begin{sourceequation}{R+}\label{eq:spectral-forced}
\begin{aligned}
a_{k,t+1}&=(1-\eta\lambda_k)a_{k,t}+b_{k,t},\\
b_{k,t}&=-\eta v_k^\top(K_t-K_0)r_t+v_k^\top\varepsilon_t.
\end{aligned}
\end{sourceequation}
Writing $\rho_k=1-\eta\lambda_k$, the exact trajectory is
\begin{sourceequation}{R+}\label{eq:spectral-history}
a_{k,t}=\rho_k^t a_{k,0}+\sum_{s=0}^{t-1}\rho_k^{t-1-s}b_{k,s}.
\end{sourceequation}
If $|\rho_k|<1$ and $|b_{k,s}|\leq B_k$, then
\begin{sourceequation}{R+}\label{eq:spectral-envelope}
|a_{k,t}|\leq|\rho_k|^t|a_{k,0}|+B_k\frac{1-|\rho_k|^t}{1-|\rho_k|}.
\end{sourceequation}
For the modal capability $C_{k,t}=\mathbf1[|a_{k,t}|<\tau_k]$, loss occurs exactly when
\begin{sourceequation}{R+}\label{eq:spectral-capability}
|\rho_k a_{k,t}+b_{k,t}|\geq\tau_k.
\end{sourceequation}
Thus the forcing history controls preservation and reacquisition in an evolving kernel.

For $f(a,b)=(2a,b+a^2)^\top$, $y=(-1,0)^\top$, squared loss, $(a,b)=(0,0)$, and $\eta=0.2$, the actual update is $(-0.4,0)$. The fixed-kernel prediction for the residual is $(0.2,0)^\top$, while the realized residual is $(0.2,0.16)^\top$. At tolerance $\tau_2=0.1$, the second modal capability is lost through the finite-amplitude forcing term. Experimental connections: TL-E03--04 and TL-E06.

\subsection{Dynamical Isometry: finite layer interactions and capability stability}\label{sec:ex:isometry}
\textbf{TLI equations \Tmark.} The exact parameterized form of the functional response~\eqref{eq:response} is:
\begin{sourceequation}{T}\label{eq:isometry-T}
R_q(\theta,\delta\theta)=m_q(\theta+\delta\theta)-m_q(\theta).
\end{sourceequation}
For differentiable readouts and sufficiently small $\delta\theta$, this has the local first-order form $R_q(\theta,\delta\theta)\simeq\nabla_\theta m_q^\top\delta\theta$.

\textbf{Original mathematical conditions \Amark.} Use the original differentiable layered computation $h^{\ell+1}=F_\ell(h^\ell,\theta_\ell)$ and layer-to-output Jacobians studied in dynamical isometry \cite{pennington2017}. Write $B_\ell=\partial F_\ell/\partial\theta_\ell$ and $J_{L\leftarrow\ell+1}=\partial h^L/\partial h^{\ell+1}$.

\textbf{Substitution and recovery \Rmark.} By the chain rule, a small parameter action in layer $\ell$ gives $\delta h^{\ell+1}\simeq B_\ell\delta\theta_\ell$, which the subsequent layers propagate. Summing across layers gives
\begin{sourceequation}{R}\label{eq:isometry-output}
\delta f\simeq \sum_\ell J_{L\leftarrow\ell+1}B_\ell\delta\theta_\ell.
\end{sourceequation}
For an activation perturbation $v$ propagated through a Jacobian $J$,
\begin{sourceequation}{R}\label{eq:isometry-sing}
\sigma_{\min}^{\mathrm{all}}(J)\lVert v\rVert \le\lVert Jv\rVert\le\sigma_{\max}(J)\lVert v\rVert,\qquad \sigma_{\min}^{\mathrm{all}}(J)=\inf_{\lVert v\rVert=1}\lVert Jv\rVert.
\end{sourceequation}
If $J$ has a nonzero kernel, the lower bound on the whole input space is zero; a positive smallest nonzero singular value bounds directions in the orthogonal complement of that kernel. Nonzero singular values close to one preserve norms along their corresponding directions. Both the layer-to-output Jacobian and the parameter map $B_\ell$ contribute to the local response.

\textbf{Finite-layer consequence \Rplusmark.} Consider the deep linear map $f(x)=W_L\cdots W_1x$ with simultaneous increments $D_\ell$, and the margin $m_q=w_q^\top f(x)-\tau_q$. For two layers,
\begin{sourceequation}{R+}\label{eq:isometry-exact}
\Delta f=(W_2D_1+D_2W_1+D_2D_1)x.
\end{sourceequation}
For arbitrary depth, let $M_\ell(S)=D_\ell$ when $\ell\in S$ and $M_\ell(S)=W_\ell$ otherwise. Expansion in layer order gives
\begin{sourceequation}{R+}\label{eq:isometry-expansion}
\Delta f=\sum_{\varnothing\ne S\subseteq\{1,\ldots,L\}}M_L(S)\cdots M_1(S)x.
\end{sourceequation}
The $|S|=1$ terms form the first-order response $\Delta f^{(1)}$. For square orthogonal receiving weights and $d_\ell=\|D_\ell\|_2$, the remaining interaction $\varepsilon_f=\Delta f-\Delta f^{(1)}$ satisfies
\begin{sourceequation}{R+}\label{eq:isometry-bound}
\|\varepsilon_f\|_2\leq\|x\|_2\left[\prod_{\ell=1}^L(1+d_\ell)-1-\sum_{\ell=1}^Ld_\ell\right].
\end{sourceequation}
Let $b_q$ be this bound multiplied by $\|w_q\|_2$ and $\widehat m_q=m_q+w_q^\top\Delta f^{(1)}$. Then
\begin{sourceequation}{R+}\label{eq:isometry-capability}
\widehat m_q>b_q\Rightarrow C_{t+1,q}=1,\qquad
\widehat m_q\leq-b_q\Rightarrow C_{t+1,q}=0.
\end{sourceequation}

For $W_1=W_2=I_2$, $x=w_q=e_1$, $\tau_q=1/2$, and the planar rotation $Q$ by $\pi/3$, compare final weights $(Q,Q^{-1})$ and $(Q,Q)$. Both interventions have identical layerwise increment norms and all initial and final input--output singular values equal to one. Their margins are $1/2$ and $-1$, respectively. Finite layer interactions therefore determine opposite capability outcomes under identical isometry and update-norm measurements. Experimental connections: TL-E03--04 and TL-E05--07.

\subsection{Maximal Update Parameterization ($\mu$P): finite feature and output responses at growing width}\label{sec:ex:mup}
\textbf{TLI equations \Tmark.} The realized parameter transition and target response in equations~\eqref{eq:update} and \eqref{eq:response} are:
\begin{sourceequation}{T}\label{eq:mup-T}
\theta_{t+1}=\theta_t+\delta_t,\qquad R_{t,q}=m_q(L_{t+1})-m_q(L_t).
\end{sourceequation}
These relations track the functional response at every parameter-increment scale.

\textbf{Original mathematical conditions \Amark.} Use the original two-hidden-layer linear MLP and SGD width-scaling illustration in Appendix E of Tensor Programs IV \cite{yang2021}. The hidden width is $n\to\infty$, with $h=U\xi$, $\overline h=Wh$, $h^\top h=\Theta(n)$, Gaussian initialization at the original $n^{-1/2}$ scale for the relevant intermediate/output weights, output $f=n^{-1/2}v\overline h$, nondegenerate loss derivative $\chi=\mathcal L'(f,y)=\Theta(1)$, and learning rate $\eta_n\propto n^{-c}$.

\textbf{Substitution and recovery \Rmark.} For one realized SGD update of the intermediate layer, the original chain-rule calculation gives
\begin{sourceequation}{A}\label{eq:mup-step}
\Delta\overline h=-n^{-c-1/2}\chi\,v^\top(h^\top h),\qquad \Delta f=-n^{-c-1}\chi\,(vv^\top)(h^\top h).
\end{sourceequation}
With $v_j=\Theta(n^{-1/2})$, $vv^\top=\Theta(1)$ and $h^\top h=\Theta(n)$ in this particular source parameterization,
\begin{sourceequation}{R}\label{eq:mup-scaling}
\Delta\overline h_j=\Theta(n^{-c}),\qquad\Delta f=\Theta(n^{-c}).
\end{sourceequation}
For $c>0$, these one-step responses vanish with width, whereas $c=0$ yields $\Theta(1)$ feature and output responses. The source appendix uses appropriately scaled output weights so the updated-feature contribution remains $\Theta(1)$, and its first-layer modification extends nondegenerate update scaling to the input layer \cite{yang2021}.

\textbf{Interaction consequence \Rplusmark.} Consider the width-$n$ linear feature model $h(q)=Uq$, $f(q)=n^{-1}v^\top h(q)$, with fixed input dimension. For a scalar loss at input $x$ with output derivative $g$, use learning rates $n\eta_h$ for $U$ and $n\eta_v$ for $v$. Then
\begin{sourceequation}{A}\label{eq:mup-two-updates}
\Delta U=-\eta_hgvx^\top,\qquad
\Delta v=-\eta_vgh(x),\qquad
\Delta h(q)=-\eta_hgv(x^\top q).
\end{sourceequation}
With $\langle a,b\rangle_n=a^\top b/n$, simultaneous substitution of both updates gives
\begin{sourceequation}{R+}\label{eq:mup-exact-response}
\begin{aligned}
R_q={}&-\eta_vg\langle h(x),h(q)\rangle_n
-\eta_hg(x^\top q)\langle v,v\rangle_n\\
&+\eta_h\eta_vg^2(x^\top q)f(x),\\
C_{t+1,q}={}&\mathbf1[m_q+R_q>0].
\end{aligned}
\end{sourceequation}
The final term is the exact interaction between feature and output-weight changes. With $v=h(x)=\mathbf1_n$, $x=q=e_1$, $g=1$, and $\eta_h=\eta_v=1/2$, the two first-order responses are $-1/2$ each and the interaction is $1/4$ at every width. The final output is $1/4$, so the interaction remains finite as $n\to\infty$.

For even $n$, let $x=e_1$, $q_A=e_2$, $q_B=e_3$, $v=\mathbf1_n$, and $s=(2,-2,\ldots,2,-2)^\top$. Set the first three columns of $U$ to $s$, $\mathbf1_n$, and $\mathbf1_n+s/2$. Under squared loss at $x$ with target $-1$, take $g=1$ and $\eta_h=\eta_v=1/2$. The two query outputs become $1$ and $0$ at every even width. At threshold $1/2$, one initially acquired capability persists and the other disappears. Width scaling preserves a target-specific finite response. Experimental connections: TL-E03--04 and TL-E05--07.

\subsection{Gradient Noise Scale: target-specific functional drift}\label{sec:ex:gns}
\textbf{TLI equations \Tmark.} The realized receiving-state update and target response~\eqref{eq:update}--\eqref{eq:response} are:
\begin{sourceequation}{T}\label{eq:gns-T}
\delta_t=\Psi(\theta_t,\omega_t,B_t),\qquad \theta_{t+1}=\theta_t+\delta_t,\qquad R_{t,q}=m_q(L_{t+1})-m_q(L_t).
\end{sourceequation}

\textbf{Original mathematical conditions \Amark.} Use independently sampled per-example gradients of \cite{mccandlish2018}, with $\mathbb E[g_i]=g$, $\Cov(g_i)=\Sigma_g$, batch size $B$, ordinary SGD and small-step linearization for the eventual functional response. Here $B$ denotes minibatch size. The formed update is
\begin{sourceequation}{A}\label{eq:gns-update}
\widehat g_B=\frac1B\sum_{i=1}^B g_i,\qquad\delta=-\eta\widehat g_B.
\end{sourceequation}
\textbf{Substitution and recovery \Rmark.} Independence yields $\Cov(\widehat g_B)=\Sigma_g/B$, hence the \emph{exact} parameter-action covariance
\begin{sourceequation}{R}\label{eq:gns-cov}
\Cov(\delta)=\frac{\eta^2}{B}\Sigma_g.
\end{sourceequation}
For a registered target $q$, let $j_q=\nabla_\theta m_q(\theta)$. The local first-order specialization of TLI response gives $R_q(\theta,\delta)\simeq j_q^\top\delta$, and therefore
\begin{sourceequation}{R}\label{eq:gns-var}
\mathbb E[R_q]\simeq-\eta j_q^\top g,\qquad \Var(R_q)\simeq\frac{\eta^2}{B}j_q^\top\Sigma_gj_q.
\end{sourceequation}
For the full output Jacobian $J$, the corresponding response covariance is
\begin{sourceequation}{R}\label{eq:gns-outputcov}
\Cov(\Delta f)\simeq\frac{\eta^2}{B}J\Sigma_gJ^\top.
\end{sourceequation}
This maps parameter update noise into target-dependent functional noise using the current receiving state's readout derivative. The source gradient noise scale is $\tr(\Sigma_g)/\|g\|^2$.

\textbf{Curvature-induced drift \Rplusmark.} Write $\delta=\mu+\xi$ with $\mu=-\eta g$ and $\mathbb E[\xi]=0$. The functional effect of stochasticity relative to the mean update is
\begin{sourceequation}{R+}\label{eq:gns-drift}
D_q=\mathbb E[R_q(\theta,\mu+\xi)]-R_q(\theta,\mu).
\end{sourceequation}
If $m_q$ is twice continuously differentiable on the noise segments, zero mean cancels the first-order term and
\begin{sourceequation}{R+}\label{eq:gns-curvature}
D_q=\mathbb E\int_0^1(1-s)\xi^\top
H_{m_q}(\theta+\mu+s\xi)\xi\,ds.
\end{sourceequation}
For small noise this becomes
\begin{sourceequation}{R+}\label{eq:gns-second}
D_q\simeq\frac{\eta^2}{2B}\tr[H_{m_q}(\theta+\mu)\Sigma_g],
\end{sourceequation}
and the relation is exact for quadratic margins.

Let $m_q(a,b)=ab-0.98$ at $(a,b)=(1,1)$ with $\mu=0$. Compare equiprobable noise $\xi=\pm(0.2,0.2)$ with $\xi=\pm(0.2,-0.2)$. Both distributions have zero mean and covariance trace $0.08$. Their mean functional responses are $0.04$ and $-0.04$, and their capability-loss probabilities are $1/2$ and $1$. Equal scalar noise magnitude can therefore produce opposite directed drift and different boundary-crossing probabilities. Experimental connections: TL-E03--04 and TL-E05--07.

\subsection{Double Descent: finite-time functional risk and early stopping}\label{sec:ex:double}
\textbf{TLI equations \Tmark.} The parameter transition and test-risk response from equations~\eqref{eq:update} and \eqref{eq:response} are:
\begin{sourceequation}{T}\label{eq:double-T}
\theta_{t+1}=\theta_t+\delta_t,\qquad R_{t,\mathrm{test}}=m_{\mathrm{test}}(L_{t+1})-m_{\mathrm{test}}(L_t).
\end{sourceequation}
In this specialization $m_{\mathrm{test}}=-\mathcal E_{\mathrm{test}}$.

\textbf{Original mathematical conditions \Amark.} Use the isotropic, zero-signal Gaussian ridgeless linear-regression model of \cite{hastie2022}: $X\in\R^{n\times p}$ with $X_{ij}\stackrel{\mathrm{iid}}\sim\mathcal N(0,1)$, $y=\varepsilon\sim\mathcal N(0,\sigma^2I_n)$, independent test covariate $x\sim\mathcal N(0,I_p)$, zero initialization, squared-loss gradient descent with the original convergence conditions, ordinary least squares for $p<n$ and minimum-norm interpolation for $p>n$.

\textbf{Substitution and recovery \Rmark.} The source action is $\delta_t=-\eta X^\top(X\beta_t-y)$. Every update lies in the row space of $X$. Inserting this action into the TLI parameter transition and taking the converged state gives
\begin{sourceequation}{R}\label{eq:dd-solutions}
\widehat\beta=\begin{cases}
(X^\top X)^{-1}X^\top y,&p<n,\\
X^\top(XX^\top)^{-1}y,&p>n.
\end{cases}
\end{sourceequation}
The original population excess prediction risk is $\mathcal E_{\mathrm{test}}(\beta)=\mathbb E_x(x^\top\beta)^2=\lVert\beta\rVert^2$, so the TLI functional response is exactly the negative stepwise change in this readout. Evaluating the final trained state and averaging over the Gaussian noise gives
\begin{sourceequation}{R}\label{eq:dd-traces}
\mathbb E[\mathcal E_{\mathrm{test}}]=\sigma^2\begin{cases}
\mathbb E_X\tr[(X^\top X)^{-1}],&p<n,\\
\mathbb E_X\tr[(XX^\top)^{-1}],&p>n.
\end{cases}
\end{sourceequation}
The inverse-Wishart expectations of the original independent Gaussian design satisfy, in their admissible parameter ranges,
\begin{sourceequation}{R}\label{eq:dd-wishart}
\mathbb E_X[(X^\top X)^{-1}]=\frac{I_p}{n-p-1},\quad p<n-1,\qquad \mathbb E_X[(XX^\top)^{-1}]=\frac{I_n}{p-n-1},\quad p>n+1.
\end{sourceequation}
Taking traces recovers
\begin{sourceequation}{R}\label{eq:dd-risk}
\mathbb E[\mathcal E_{\mathrm{test}}]=\begin{cases}
\displaystyle\sigma^2\frac{p}{n-p-1},&p<n-1,\\[4pt]
\displaystyle\sigma^2\frac{n}{p-n-1},&p>n+1.
\end{cases}
\end{sourceequation}
This finite-sample noise excess risk rises toward the interpolation region and decreases on the overparameterized side under the stated Gaussian conditions.

\textbf{Finite-time consequence \Rplusmark.} Conditional on a realized design, take $0<\eta<2/\lambda_{\max}(X^\top X)$ and let $\lambda_i$ be its positive eigenvalues with $a_i=1-\eta\lambda_i$. The row-space solution after $t$ steps is
\begin{sourceequation}{R+}\label{eq:dd-path}
\theta_t=\sum_{k=0}^{t-1}(I-\eta X^\top X)^k\eta X^\top y.
\end{sourceequation}
The conditional noise-averaged excess test risk is
\begin{sourceequation}{R+}\label{eq:dd-finite-risk}
\overline{\mathcal E}_t=\mathbb E_y[\|\theta_t\|^2\mid X]
=\sigma^2\sum_{i=1}^{\operatorname{rank}(X)}\frac{(1-a_i^t)^2}{\lambda_i}.
\end{sourceequation}
This is the finite-time spectral filter associated with iterative regularization \cite{raskutti2014early}. Since $m_{\rm test}=-\mathcal E_{\rm test}$, the stepwise TLI response is
\begin{sourceequation}{R+}\label{eq:dd-response}
\mathbb E_y[R_{t,\mathrm{test}}\mid X]
=-\eta\sigma^2\sum_i a_i^t[2-(1+a_i)a_i^t].
\end{sourceequation}
When $0<\eta\lambda_i\leq1$ for every positive mode, each response is nonpositive: training progressively accumulates fitted noise while training loss falls. For fixed $t$ and small $\lambda_i$, a mode contributes $\sigma^2t^2\eta^2\lambda_i+o(\lambda_i)$, compared with the converged contribution $\sigma^2/\lambda_i$. Early stopping therefore truncates the modes that produce the interpolation peak.

For a risk budget $\tau>0$, define $m_\tau=\tau-\overline{\mathcal E}_t$ and $C_{t,\tau}=\mathbf1[\overline{\mathcal E}_t<\tau]$. Its exact cumulative mean-response boundary is
\begin{sourceequation}{R+}\label{eq:dd-budget}
C_{t,\tau}=0\quad\Longleftrightarrow\quad
\sum_{k=0}^{t-1}\mathbb E_y[R_{k,\mathrm{test}}\mid X]\leq-\tau.
\end{sourceequation}
The converged limit recovers equation~\eqref{eq:dd-risk}; the finite trajectory supplies a capability boundary for the early-stopping risk budget. Experimental connections: TL-E03 and TL-E06--07.

\subsection{Summary of the ten specializations}\label{sec:ex:summary}
The ten cases use the same TLI receiving-state, actual-action, functional-response, capability-readout, and feedback relations with the respective source-model conditions. They recover established relations in Edge of Stability, catastrophic forgetting, task arithmetic, Neural Tangent Kernel theory, recursive model collapse, Gradient Noise Scale, Spectral Bias, maximal update parameterization, dynamical isometry, and double descent. The \Rplusmark{} results extend the same substitutions to optimizer-memory-dependent capability boundaries, finite-amplitude interactions, stochastic functional drift, forced modal dynamics, finite-time risk trajectories, and feedback-allocation boundaries.

\clearpage
\section{Equation-organized experimental ledgers}\label{app:experimental-displays}
The following ledgers unfold the coupled dynamics in experimental order. They
separate action formation, receiving-state response, persistent learning,
frozen inference, feedback, and quantitative boundary prediction. A dash marks
a quantity that was not reported under a directly comparable protocol; it does
not denote a zero. Registered failures remain visible.

\begin{figure}[!htbp]
\centering\includegraphics[width=.80\textwidth]{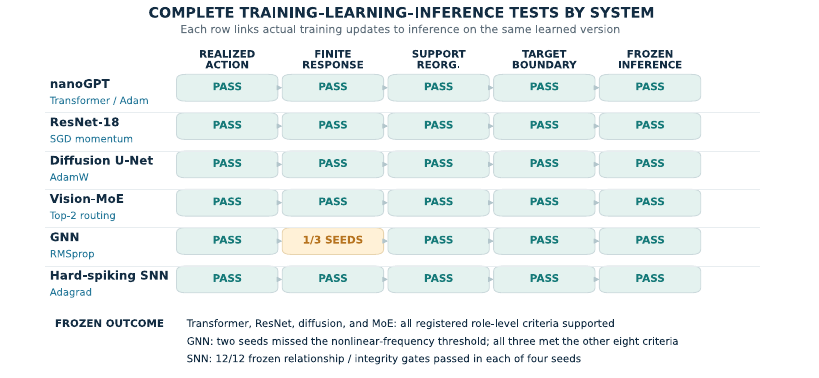}
\caption{Within-system training--learning--inference experiment chains. Two
GNN seeds meet eight of nine registered criteria and miss the nonlinear-response
frequency threshold. The third GNN seed meets all nine. The hard-spiking SNN
meets all twelve structural criteria in four seeds; its separate boundary
predictor does not meet the registered prediction criterion.}
\label{fig:cross}
\end{figure}
\FloatBarrier

\begin{table}[!htbp]
\centering
\begin{minipage}{0.96\textwidth}
\caption{Training state and realized-action formation. This ledger instantiates
Eqs.~\eqref{eq:training}--\eqref{eq:update} and the optimizer-memory intervention
in Eq.~\eqref{eq:exp2}. Formula errors are maximum absolute reconstruction
errors; endpoint errors are shown only where the same frozen summary reports
them.}\label{tab:training-state}\label{tab:systems}
\footnotesize
\renewcommand{\arraystretch}{1.15}
\begin{tabularx}{\linewidth}{@{}l l >{\raggedright\arraybackslash}X >{\raggedright\arraybackslash}X r@{}}
\toprule
System & Optimizer & Registered state/action records & Optimizer-memory test & Max error \\
\midrule
nanoGPT & Adam & 13 frozen histories; 3 replayable checkpoint/GFG histories & Optimizer-separated causal forks; replay checks PASS & -- \\
ResNet-18 & SGD momentum & 1,536 target-update records; $\alpha=0/1$ errors 0 & 18/21 exchanges above threshold & $5.96\!\times\!10^{-8}$ \\
Diffusion U-Net & AdamW & 504 target-update records; $\alpha=0/1$ errors 0 & 18/18 exchanges above threshold & $1.49\!\times\!10^{-8}$ \\
Vision-MoE & AdamW & 1,344 target-update records; $\alpha=0/1$ errors 0 & 18 registered exchanges; frozen test PASS & $1.19\!\times\!10^{-7}$ \\
GNN & RMSprop & 12 hash-identified pre/update/post states in 3 seeds & Not separately summarized in the frozen formal result & -- \\
Hard-spiking SNN & Adagrad & Native states in 4 seeds & 12 exchanges; 4/4 seed gates PASS & $5.02\!\times\!10^{-8}$ \\
\bottomrule
\end{tabularx}
\end{minipage}
\end{table}
\FloatBarrier

\begin{table}[!htbp]
\centering
\begin{minipage}{0.96\textwidth}
\caption{Receiving-state dependence and finite-amplitude response. The same
formed update is evaluated in different receiving states
(Eqs.~\eqref{eq:response} and \eqref{eq:exp1}); complete recorded updates are
then scaled along Eq.~\eqref{eq:exp3}. ``Nonlinear'' uses each frozen protocol's
registered threshold.}\label{tab:response}
\footnotesize
\renewcommand{\arraystretch}{1.15}
\begin{tabularx}{\linewidth}{@{}l >{\raggedright\arraybackslash}X r r r >{\raggedright\arraybackslash}X@{}}
\toprule
System & Receiving-state intervention & Finite-response paths & Turnback & Sign reversal & Registered result \\
\midrule
nanoGPT & Reciprocal matched-pair response changed in both directions & 15,264; five morphologies & 5,510 & 103 & SUPPORTED \\
ResNet-18 & 42/42 exchanges above threshold & 495/1,536 (32.23\%) & 80 & 143 & SUPPORTED \\
Diffusion U-Net & 18/18 exchanges above threshold & 97/504 (19.25\%) & 9 & 16 & SUPPORTED \\
Vision-MoE & 36/36 exchanges; frozen test PASS & 480/1,344 (35.71\%) & 53 & 76 & SUPPORTED \\
GNN & 9 exchanges; same-update gate PASS in 3/3 seeds & 62/768 (8.07\%) & -- & -- & 10\% frequency gate: 1/3 seeds \\
Hard-spiking SNN & 12 exchanges; same-update gate PASS in 4/4 seeds & 718/1,536 (46.74\%) & -- & -- & SUPPORTED in 4/4 seeds \\
\bottomrule
\end{tabularx}
\end{minipage}
\end{table}
\FloatBarrier

\begin{table}[!htbp]
\centering
\begin{minipage}{0.96\textwidth}
\caption{Distributed support and non-additive component interaction. This
ledger instantiates the support object in Eq.~\eqref{eq:support}, the measured
interaction in Eq.~\eqref{eq:exp6}, and the four-component coalition protocol
in Eq.~\eqref{eq:exp9}. Percentages are within-system distributed-support
detection rates.}\label{tab:support-ledger}
\footnotesize
\renewcommand{\arraystretch}{1.15}
\begin{tabularx}{\linewidth}{@{}l >{\raggedright\arraybackslash}X r >{\raggedright\arraybackslash}X@{}}
\toprule
System & Registered intervention unit & Distributed support & Non-additive evidence \\
\midrule
nanoGPT & 16 coalitions over 4 attention/MLP branches & Witnessed & $\geq138$ non-additive pair-groups per audited phase \\
ResNet-18 & 16 coalitions over 4 residual stages & 98.63\% & SUPPORTED \\
Diffusion U-Net & 16 coalitions over 4 U-Net paths & 76.59\% & SUPPORTED \\
Vision-MoE & 16 coalitions over 4 routed experts & 94.72\% & SUPPORTED \\
GNN & 4 message-passing blocks & 3/3 seeds & 3/3 seeds \\
Hard-spiking SNN & 16 coalitions over 4 spiking blocks & 4/4 seeds & 4/4 seeds \\
\bottomrule
\end{tabularx}
\end{minipage}
\end{table}
\FloatBarrier

\begin{table}[!htbp]
\centering
\begin{minipage}{0.96\textwidth}
\caption{Persistent support reorganization and capability-boundary readout.
This ledger connects $\Delta P$ in Eqs.~\eqref{eq:support} and
\eqref{eq:exp4} to $C_{t+1,q}$ in Eq.~\eqref{eq:capability}. The original
frozen nanoGPT corpus and the later replayable supplement are reported as
separate evidence lineages.}\label{tab:learning-boundary}
\footnotesize
\renewcommand{\arraystretch}{1.15}
\begin{tabularx}{\linewidth}{@{}l r r r >{\raggedright\arraybackslash}X@{}}
\toprule
System / evidence lineage & Comparisons & Reallocation & Primary handoff & Capability relation \\
\midrule
nanoGPT, original frozen & 1,636 & -- & 228 (13.94\%) & Frozen witnesses include $\Delta P\ne0$ while $\Delta C=0$ \\
nanoGPT, replay supplement & 828 & mean 0.0356 & 24 (2.90\%) & 617 unchanged: 0.0153 mean; 211 changed: 0.0950 mean \\
ResNet-18 & 1,536 & 91.41\% & 8.85\% & Identity-aligned endpoint boundaries \\
Diffusion U-Net & 504 & 97.02\% & 8.13\% & Identity-aligned endpoint boundaries \\
Vision-MoE & 1,344 & 80.43\% & 9.97\% & Identity-aligned endpoint boundaries \\
GNN & 3 seeds & 3/3 seeds & -- & Persistent-reallocation criterion met \\
Hard-spiking SNN & 4 seeds & 4/4 seeds & -- & Persistent-reorganization criterion met \\
\bottomrule
\end{tabularx}
\end{minipage}
\end{table}
\FloatBarrier

\begin{table}[!htbp]
\centering
\begin{minipage}{0.96\textwidth}
\caption{Frozen inference, query-conditioned recruitment, non-additive
combination, and learned-version rollback. This ledger tests
Eq.~\eqref{eq:inference} through Eqs.~\eqref{eq:exp5}--\eqref{eq:exp7}.
Interaction and rollback magnitudes use native system readout scales and must
not be ranked across rows.}\label{tab:inference-ledger}
\footnotesize
\renewcommand{\arraystretch}{1.15}
\begin{tabularx}{\linewidth}{@{}l >{\raggedright\arraybackslash}X r r r l@{}}
\toprule
System & Frozen execution & Query separation & Pair interaction & Rollback effect & Restore \\
\midrule
nanoGPT & 52 phase-specific comparisons & min 0.0632 & max 6.5402 & changed in 52 phases & exact \\
ResNet-18 & 3/3 seeds; repeat drift 0 & max 2.0000 & max 32.5550 & max 32.0113 & 0 \\
Diffusion U-Net & 3/3 seeds; repeat drift 0 & max 2.0000 & max $6.54\!\times\!10^{-4}$ & max 3.1794 & 0 \\
Vision-MoE & 3/3 seeds; model/optimizer hashes fixed & max 1.7946 & max 6.6726 & max 10.3805 & 0 \\
GNN & 3/3 seeds; frozen execution exact & max 2.0000 & max 7.9576 & max 15.2035 & exact \\
Hard-spiking SNN & 4/4 seeds; frozen execution exact & max 2.0000 & max 3.9745 & max 10.8089 & exact \\
\bottomrule
\end{tabularx}
\end{minipage}
\end{table}
\FloatBarrier

\begin{table}[!htbp]
\centering
\begin{minipage}{0.96\textwidth}
\caption{Feedback-directed support reorganization and recovery, closing the
loop in Eq.~\eqref{eq:feedback}. RL-E05's registered overall verdict remains
NOT SUPPORTED because only 2/12 seeds met its temporal-precedence gate; RL-E06
then tests dose ordering and exact-state recovery forks.}\label{tab:feedback-ledger}
\footnotesize
\renewcommand{\arraystretch}{1.15}
\begin{tabularx}{\linewidth}{@{}l >{\raggedright\arraybackslash}X r l@{}}
\toprule
\multicolumn{4}{@{}l}{\emph{Panel A: RL-E05 registered gates}}\\
Gate & Meaning & Seeds met & Overall verdict \\
\midrule
Directional rollback & Component-version intervention moves the registered trade-off in the predicted direction & 12/12 & partial \\
New-component capture & Selective feedback recruits a newly positive component & 11/12 & partial \\
Temporal precedence & Required support-share increase precedes the first registered accuracy loss & 2/12 & NOT SUPPORTED \\
\bottomrule
\end{tabularx}

\vspace{0.8em}
\begin{tabular*}{\linewidth}{@{\extracolsep{\fill}}lrrl@{}}
\toprule
\multicolumn{4}{@{}l}{\emph{Panel B: RL-E06 dose allocation; all 12 seeds follow both registered orderings}}\\
Allocation per update & Skill 0 support share & Other-skill mean margin & Verdict \\
\midrule
Balanced $(15,15,15,15)$ & 25.54\% & 9.474 & SUPPORTED \\
Mild $(30,10,10,10)$ & 26.47\% & 9.130 & SUPPORTED \\
High $(45,5,5,5)$ & 27.50\% & 8.493 & SUPPORTED \\
Exclusive $(60,0,0,0)$ & 35.20\% & 3.962 & SUPPORTED \\
\bottomrule
\end{tabular*}

\vspace{0.8em}
\begin{tabularx}{\linewidth}{@{}l >{\raggedright\arraybackslash}X r r@{}}
\toprule
\multicolumn{4}{@{}l}{\emph{Panel C: exact-state fork and recovery}}\\
Branch & Intervention & Other-skill accuracy & Skill 0 accuracy \\
\midrule
Common fork, update 800 & Identical parameter--AdamW state before recovery branching & 70.31\% & -- \\
Continue exclusive, update 3200 & Continue concentrated feedback & 60.94\% & -- \\
Rebalanced recovery & Restore balanced allocation & 99.48\% & 100.00\% \\
Other-skills repair & Train only the other three skills & 99.83\% & 89.06\% \\
\bottomrule
\end{tabularx}
\end{minipage}
\end{table}
\FloatBarrier

\begin{table}[!htbp]
\centering
\begin{minipage}{0.96\textwidth}
\caption{Post-update target-boundary prediction from the formed update and
receiving-state response (Eqs.~\eqref{eq:update}, \eqref{eq:capability}, and
\eqref{eq:exp8}). Panel A reports four-way macro recall (MR) and the selected
predictor's endpoint accuracy. Panel B is a separate run-isolated protocol; its
values must not be compared numerically with Panel A as if they shared a split.}
\label{tab:pred}
\footnotesize
\renewcommand{\arraystretch}{1.15}
\begin{tabular*}{\linewidth}{@{\extracolsep{\fill}}lrrrrrl@{}}
\toprule
\multicolumn{7}{@{}l}{\emph{Panel A: direct endpoint prediction}}\\
System & Targets & No-change MR & First-order MR & First+second MR & Accuracy & Status \\
\midrule
nanoGPT (confirmation) & 5,088 & 50.00 & 83.81 & 91.49 & 91.43 & SUPPORTED \\
ResNet-18 & 1,536 & 50.00 & 97.64 & 97.99 & 99.28 & SUPPORTED \\
Diffusion U-Net & 504 & 50.00 & 98.96 & 97.98 & 99.60 & SUPPORTED \\
Vision-MoE & 1,344 & 50.00 & 96.93 & 96.95 & 99.63 & SUPPORTED \\
GNN & 3,072 & 50.00 & 94.68 & 97.87 & 99.87 & SUPPORTED \\
Hard-spiking SNN & 7,200 & 50.00 & 60.05 & 60.05 & 93.24 & NOT SUPPORTED \\
\bottomrule
\end{tabular*}

\vspace{0.8em}
\begin{tabular*}{\linewidth}{@{\extracolsep{\fill}}lrrrr@{}}
\toprule
\multicolumn{5}{@{}l}{\emph{Panel B: run-isolated pairing controls (four-way MR, \%)}}\\
System & Correct pairing & Target permuted & Update geometry permuted & Optimizer memory permuted \\
\midrule
ResNet-18 & 80.06 & 57.95 & 59.49 & 74.99 \\
Diffusion U-Net & 63.59 & 56.84 & 51.02 & 60.84 \\
Vision-MoE & 74.79 & 49.84 & 55.48 & 66.45 \\
\bottomrule
\end{tabular*}
\end{minipage}
\end{table}
\FloatBarrier

{\raggedright
Machine-readable values and package-relative evidence routes for
Tables~\ref{tab:training-state}--\ref{tab:pred} are in
\texttt{tables/APPENDIX\_EQUATION\_TABLES.json}. Per-seed records,
checkpoints, and intervention outputs remain in the accompanying evidence
package.\par}

\clearpage
\onecolumn
\section{Research materials and disclosures}\label{app:disclosures}
\subsection*{Data, code, and materials availability}
A complete evidence package accompanies this preprint. It contains the frozen experimental source code and protocols, machine-readable results, model checkpoints, captured Generation-Fact Graph records, mathematical audit materials, and executable validation and replay procedures.

A claim-to-evidence matrix links every central empirical claim to its protocol, result, source code, checkpoint or runtime artifact, and corresponding checker. The package separates direct result recomputation, checkpoint-based replay, and optional fresh training, and provides SHA-256 manifests for both the compact reviewer package and the full evidence archive.

\subsection*{Funding and competing interests}
This research received no funding. The author declares no competing interests.

\subsection*{Author contributions}
The author conceived the study; developed the Generation-Fact Graph, the GFG-based recursive scientific process, and the unified theory of training, learning, and inference; defined the scientific questions, research boundaries, and decision criteria; introduced the principal theoretical hypotheses and conceptual transitions; directed the research program; analyzed and interpreted validated results; reviewed and selected the cross-theory extrapolations produced primarily by the AI scientific agents; prepared the figures; and wrote and revised the manuscript.

\end{document}